\documentclass[
  journal=chr,
  manuscript=article,
  hyperref,
]{cup-journal}

\usepackage[utf8]{inputenc}
\DeclareUnicodeCharacter{1E25}{\d{h}}%
\DeclareUnicodeCharacter{1E24}{\d{H}}
\usepackage{amsmath}
\usepackage{microtype}
\usepackage{booktabs}
\usepackage{array}
\usepackage{xcolor}
\makeatletter\@ifundefined{KV@Gin@alt}{\define@key{Gin}{alt}{}}{}\makeatother

\usepackage{algorithm}
\usepackage{algpseudocode}

\graphicspath{{Fig/}}

\makeatletter\@twosidefalse\@mparswitchfalse\makeatother
\AtBeginDocument{\hypersetup{bookmarksdepth=2}}

\newif\ifanon\anonfalse

\title{Selective Reception: Tracing How Super-Commentators Read Nachmanides' Secrets with Validated Knowledge Graphs}

\ifanon
\author{Anonymous authors}
\affiliation{Withheld for anonymous review}
\else
\author{Hadar Miller}
\affiliation{Information Systems Department, University of Haifa, 199 Aba Khoushy Ave., Haifa 3498838, Israel}
\email[Hadar Miller]{miller.hadar@gmail.com}

\author{Eliezer Baumgarten}
\affiliation{Department of Jewish History, University of Haifa, 199 Aba Khoushy Ave., Haifa 3498838, Israel}

\author{Oded Yisraeli}
\affiliation{Department of Jewish Thought, Ben-Gurion University of the Negev, Marcus Family Campus, Beer-Sheva 8410501, Israel}

\author{Moshe Lavee}
\affiliation{Department of Jewish History, University of Haifa, 199 Aba Khoushy Ave., Haifa 3498838, Israel}
\fi


\abbreviations{KG (knowledge graph), LLM (large language model), PABAK (prevalence- and bias-adjusted kappa), PSA (positive specific agreement), RP (Ramban passage), SCT (super-commentary text)}

\begin{document}

\begin{abstract}
How did later kabbalists receive the commentary of Nachmanides (the Ramban, d.~1270) on the Pentateuch, and especially its allusive ``Secrets''? Their reception is largely conceptual: super-commentators take up his ideas without repeating his words, so methods that trace text reuse or citation capture only part of it. We represent 385 aligned passages, 38 comments of the Ramban and the responses to them in twenty super-commentary texts, as knowledge graphs of typed concepts and relations. The graphs are generated by large language models (LLMs) under a scholar-designed ontology, and every element is linked to the words of the source that ground it. To test how far they can be trusted, an LLM judge from a different provider scores them, two experts in the tradition check the judge, and four generators and two general-purpose tools are compared. The experts themselves agree only weakly, so the scores are read as relative; held-out $F_1$ ranges from 0.74 to 0.92, and errors concentrate in how interpretive relations are labelled rather than in invented ones. Across the corpus, conceptual comparison shows selective reception: later commentators develop particular kabbalistic hints while engaging with the Ramban's grammatical and cross-referential observations in different ways. Close reading shows that shared concepts can serve different purposes, and different vocabularies a common reading. For most passages his text remains a common anchor from which the responses diverge; for a few, such as the Shema, they share a vocabulary his text does not make explicit, elaborating hints he leaves unexplained even where they read them differently. The substantially represented works combine his material with further sources, though not every witness has a distinct profile. The comparison also recovers known witnesses of one work and flags further manuscript units for verification. The graphs locate these patterns; reading the texts interprets them.
\end{abstract}

\section*{Plain Language Summary}
In the thirteenth century, the Spanish rabbi Moses ben Nahman, known as Nachmanides or the Ramban, wrote a commentary on the Five Books of Moses that became one of the most influential works of Jewish biblical interpretation. Some of his comments hint at mystical (kabbalistic) teachings without explaining them. These hints, known as the Ramban's Secrets, attracted later kabbalists, several of whom wrote commentaries on his commentary to explain what he meant. This study asks how those later writers read him: what they took up, what they left aside, and where they share concepts and where they differ in interpreting them.

The question is hard to answer by tracing shared words, because the later writers often take up the Ramban's ideas in their own words. Reading every passage closely is slow and makes patterns across a whole tradition hard to see. We therefore used artificial intelligence, in the form of large language models, to turn each passage into a map of the ideas it contains and the connections it makes between them. Scholars of the tradition defined the kinds of ideas and connections the maps could use, and every element of every map points back to the words in the text that it came from. We built maps for 38 comments of the Ramban and for 347 responses to them in twenty later works.

Because these maps were made by a machine, we checked how far they can be trusted. A second model from a different company checked each map against its text, and two experts in the tradition checked that second model. The experts themselves often disagreed about which connections were right, which shows how much of this material is a matter of interpretation. The machine rarely invented connections; its mistakes were mostly in how it named connections that scholars might also name differently. We therefore use the maps to compare texts, not as a final word on any of them.

The comparisons show that the later writers selected and transformed elements of the Ramban's commentary rather than simply repeating them. A grammatical observation may become the starting point for a kabbalistic exposition, and passages with different vocabularies may nevertheless share an interpretive core. For most passages, each writer took the Ramban in a different direction, so his text remained the common starting point. For a few passages, such as the Shema (``Hear, O Israel''), the later writers share concepts beyond the Ramban's explicit wording, even where they differ in how they interpret his hints. Some works show recurrent ways of building on him, for example by drawing on an early mystical book or by matching divine attributes to particular divine powers. The comparison also identified two manuscript units whose close conceptual similarity warrants a textual comparison.

The maps do not replace reading. They locate patterns that scholars can investigate, and lead back to the texts in which the patterns can be interpreted. A shared concept need not have the same explanatory function in every text.

\section{Introduction}\label{sec:intro}

 Studying the history of ideas requires identifying patterns of similarity and difference among texts, and analysing broader trends in the network of relationships within a corpus. Exegetical literature poses a particular challenge, especially when it employs complex and varied imagery and makes extensive use of figurative language. Commentators preserve some elements of the texts they interpret while modifying, expanding, or setting aside others.  To trace these developments, we need to compare each commentary with the text it interprets, examining both the concepts it takes up and the ways it develops them.  Comparing several commentaries on the same text also lets us ask whether they converge or diverge. Verbal overlap and shared citations can help us trace these relationships, but they do not capture all forms of conceptual continuity and change. How can we systematically compare the conceptual content of texts across an exegetical tradition, while retaining a basis for close reading of the ways individual commentators organise and develop that content?

The exegetical tradition associated with Nachmanides (R.~Moses b.~Nahman, hereafter the Ramban), who was active in thirteenth-century Spain, offers a case in which these challenges are particularly pronounced. The Ramban's commentary on the Pentateuch became one of the foundational works of medieval Jewish biblical interpretation \citep{yisraeli2020nachmanides,idel1998nahmanides}.  His commentary combines philological, legal (halakhic), and kabbalistic layers, often within a single comment. The passages conventionally called the ``Secrets of the Ramban'' belong primarily to the kabbalistic layer. They are embedded in the biblical commentary rather than set out separately, and they convey esoteric ideas through specialised terminology and allusive phrases such as ``by way of truth'' \citep{halbertal2006way,idel1983noKabbalisticTradition,yisraeli2020nachmanides}.  A super-commentary comments on an earlier commentary rather than directly on the biblical text; despite this secondary position, the genre permits substantial interpretive independence \citep{lawee2017genre}. Later super-commentaries devoted to interpreting the Ramban's commentary include \emph{Keter Shem Tov}, \emph{Peirush Sodot ha-Torah le-Ramban}, and \emph{Me'irat Einayim}.  These works differ in their claims to transmitted authority and in their freedom to propose new readings; they do not form a single uniform interpretive practice \citep{abrams1996orality,yisraeliForthcomingTradition,lawee2017genre}. Their authors unpack the Ramban's allusions within broader conceptual systems \citep{halbertal2006way,yisraeli2023keter,yisraeli2024meditative}.

We examine this tradition through two related questions. In the vertical comparison, we ask: which elements of a Ramban passage do later super-commentators preserve, expand, or set aside? In the horizontal comparison, we compare their different interpretations of the same Ramban passage: where do they share concepts, and where do their explanatory choices differ?  Both questions concern relationships among ideas that cannot be established from recurring words alone: what persists, what disappears, and what is reorganised as a tradition develops.
Comparing these relationships across passages also reveals recurring choices of sources, interpretive vocabulary, and explanatory relations within individual works. We distinguish conceptual overlap, the terms used to express it, and the interpretive functions those terms serve. Across the aligned corpus, such comparisons can also identify candidates for shared interpretive traditions, a phenomenon already identified in selected pairs of commentaries \citep{yisraeliForthcomingTradition}; establishing the historical relationships of any new candidates requires close reading and, where relevant, manuscript evidence.

The reception examined here is largely conceptual. A super-commentator may take up an idea of the Ramban's without repeating his words or naming him: identifying a divine attribute with a particular \emph{sefirah}, for example, can be the interpretive centre of a passage without a single quotation or citation. Methods that trace text reuse or citation therefore see only part of this reception, while close reading, which can see all of it, cannot by itself establish which patterns recur across hundreds of aligned passages. Earlier work on Aggadic literature with ACT (Allocate Connections between Texts) showed that graph-based representations capture structural relationships between texts that verbal overlap alone does not \citep{miller2025alignment,miller2026aggadic,miller2026phd}. The Ramban corpus requires a further, explicitly conceptual layer.

We use knowledge graphs (KGs) for that layer. Each passage is represented as a graph whose nodes are typed concepts (kabbalistic, halakhic, exegetical, and so on) and whose edges are labelled relations between them. A shared, scholar-designed ontology supplies the categories and relations, so that the graph of a Ramban passage can be compared with the graphs of its super-commentaries, and these with one another. This comparison identifies candidates for shared concepts, rather than proving shared interpretations or historical transmission. Every concept and relation carries the verbatim source span that grounds it, so each corpus-level pattern leads back to the passages that produced it (Section~\ref{sec:methodology}).

The paper accordingly has two aims. The first is historical: to describe how the Ramban's commentary was received across 385 aligned passages, 38 Ramban passages and their counterparts in twenty super-commentary texts, and to return from the corpus-wide patterns to the texts that explain them. The second is methodological: to show that large language models (LLMs) can generate concept graphs of this kind for medieval Rabbinic Hebrew reliably enough to support such comparison, in a setting with no labelled training data, no pretrained domain models, and no gold-standard graphs.

The graphs point to \emph{selective} reception. The later commentators neither preserve nor abandon the Ramban wholesale; they elaborate kabbalistic hints and also sometimes transform his grammatical and cross-referential observations into new explanations. At passage level, the later responses to most Ramban passages remain closer to him than to one another: he serves as a common anchor from which they develop different emphases. For a few passages, such as the Shema, they share an elaborated conceptual vocabulary not explicit in his own text, although its interpretive use can differ. At work level, the substantially represented commentaries combine his material with further sources and concepts, and most show recurrent interpretive practices, although not every witness has a distinct profile. The comparison also identifies witness relationships that invite manuscript collation.

These historical results rest on a reliability argument, which Section~\ref{sec:experiments} sets out. The graphs are generated by a pipeline, \texttt{simpleKG}, that keeps corpus-specific knowledge in replaceable prompts and in an ontology refined against validator feedback. They are scored by an LLM judge from a different provider than the generator, and the judge is in turn checked against two subject-matter experts. Across four generators, held-out $F_1$ ranges from 0.74 to 0.92; the configuration used for the full corpus reaches 0.88. We test the approach on this corpus only, and whether it transfers to other traditions remains to be shown (§\ref{subsec:disc-limits}).

The claims that follow are not of one kind, and Section~\ref{sec:results} grades them as corpus measurements, model-based comparative evidence, or interpretive hypotheses that require close reading or collation. The computation finds patterns across hundreds of passages and points to the evidence that needs explanation; it does not turn similarity into historical interpretation. Section~\ref{sec:data} introduces the tradition and the corpus, Section~\ref{sec:relatedwork} situates the study, Section~\ref{sec:methodology} describes how passages become concept graphs, Section~\ref{sec:experiments} assesses how far the graphs can be trusted, Section~\ref{sec:results} presents the findings, and Section~\ref{sec:discussion} considers what they add to the study of the Ramban's reception and to the method.

\section{The tradition and the corpus}\label{sec:data}

The historical difficulty becomes especially clear in the passages conventionally described as the Ramban's Secrets. Some comments in his Pentateuch commentary point beyond their plain sense. Introduced by formulae such as ``by way of truth'', they allude to kabbalistic teachings without spelling them out, and the Ramban presented these teachings as received tradition rather than as matter for open exposition \citep{halbertal2006way,idel1983noKabbalisticTradition}. These passages, conventionally called the Secrets of the Ramban, form a distinctive esoteric stratum within his Pentateuch commentary, characterised by allusive language and deliberate reticence \citep{yisraeli2020nachmanides}.
Their reticence invited explanation. In the generations after his death, kabbalists wrote super-commentaries devoted to them, and broader commentaries and kabbalistic works took up the same passages, each unpacking the allusions within its own conceptual system \citep{idel1998nahmanides,yisraeli2023keter,yisraeli2024meditative}.

The corpus brings together twenty such super-commentary texts (SCTs; Appendix~H). Some are works devoted to the Secrets: \emph{Keter Shem Tov} by R.~Shem Tov ibn Gaon, \emph{Be'ur Sodot ha-Ramban}, a work closely associated with the teachings of R.~Joshua ibn Shuaib but of disputed authorship, \emph{Me'irat Einayim} by R.~Isaac of Acre, and \emph{Da'at Chacham} \citep{yisraeliForthcomingTradition,abrams1996orality}. Others are broader works that comment on the same passages, among them the Torah commentaries of Rabbeinu Bahya ben Asher and R.~Menachem Recanati and the kabbalistic treatise \emph{Ma'arekhet ha-Elohut}. The corpus thus compares responses to shared Ramban passages across works of different genres and purposes, rather than treating them as a uniform literary genre. Recanati's commentary, in particular, draws upon \emph{Keter Shem Tov} but is not simply another work devoted to expounding the Ramban's Secrets \citep{idel1998recanati}. Several survive only in manuscript, including anonymous units in codices at the Escorial and the Vatican, and some works are represented by more than one manuscript witness. Witnesses that differ enough to be read as separate texts are treated as separate SCTs, which also lets the analysis test whether the graphs recognise witnesses of one work as close to each other (§\ref{subsec:res-twins}).

For each of 38 Ramban passages (RPs), each a comment on a verse or short group of verses in the digitised Chavel edition, we collected the responses to it in these works from manuscripts and printed editions. The result is a sparse alignment grid: the RPs form the rows, the Ramban and the twenty SCTs the columns, and a cell is filled where a work comments on that passage. Of the 798 possible cells, 385 are filled: the 38 RPs and 347 SCT responses. We call the unit aligned to an RP an \emph{SCT response}, and the reading it develops an \emph{interpretation}. Table~\ref{tab:glossary} glosses these and the other main terms of the paper. Twelve SCTs appear in at least ten rows and form the core on which corpus-wide claims rest. The passages were chosen where the tradition is most active, so the grid is dense in interpretation but does not exhaust the tradition.

\begin{table}[!htbp]
\centering
\small
\caption{Glossary of the main terms.}\label{tab:glossary}
\begin{tabular}{>{\raggedright\arraybackslash}p{0.22\textwidth}>{\raggedright\arraybackslash}p{0.7\textwidth}}
\toprule
\textbf{Term} & \textbf{Meaning in this paper} \\
\midrule
Ramban & R.~Moses ben Nahman (Nachmanides, d.~1270), author of the commentary on the Pentateuch studied here \\
Secrets of the Ramban & The kabbalistic passages of his commentary, often marked ``by way of truth'' (\emph{derekh ha-emet}), which allude to teachings without explaining them \\
Kabbalah & Medieval Jewish esoteric traditions concerning the divine realm and its relations to the world \\
\emph{Sefirah} (pl.\ \emph{sefirot}) & In Kabbalah, one of ten divine powers or attributes, such as Tif'eret or Atarah, whose interrelations are central to the texts studied here \\
Super-commentary & A commentary on a commentary; here, a work that explains the Ramban's commentary \\
RP (Ramban passage) & One of the 38 comments of the Ramban that anchor the corpus \\
SCT, SCT response & A text commenting on the Ramban, or a distinct manuscript witness used as a separate analytic unit; its response to a given RP (not necessarily an independent work) \\
Knowledge graph & A network whose nodes are concepts and whose edges are named relations between them \\
Ontology & The agreed set of concept types and relation names that the graphs may use \\
$F_1$ & A score from 0 to 1 combining precision (how much of what was extracted is correct) and recall (how much of what is in the text was extracted) \\
$\kappa$, PSA, PABAK & Measures of agreement between two judges, each correcting differently for agreement expected by chance \\
\bottomrule
\end{tabular}
\end{table}

\noindent\textbf{A worked textual comparison (RP~32, Deuteronomy 6:4).}
The Ramban asks why Moses says ``the Lord \emph{our} God'' here rather than the more usual ``the Lord \emph{your} God''. The following short excerpts are our translations from the project transcriptions.
\begin{quote}\small
\textbf{Ramban:} ``You should reflect on this, for Scripture changes its wording here to say `our God' rather than `your God' ... But in the declaration of unity it says `the Lord our God,' because He performed great and awesome deeds with Moses, making for Himself a glorious name.''

\textbf{Keter Shem Tov:} ``Moses himself joined them in accepting the sovereignty of Heaven. Had he said `your God,' it would have appeared as though he had excluded himself from the community.''

\textbf{Ibn Shuaib:} ``Moses had to depart here from his usual formulation in Deuteronomy, `the Lord your God,' which alludes to the final letter \emph{heh}. Here, however, he needed to unite the entire great and awesome structure in complete unity.''
\end{quote}
The three texts share a grammatical starting point but develop it differently. Keter Shem Tov foregrounds Moses' participation in Israel's confession of divine unity; Ibn Shuaib reads the wording within a sefirotic account of unification. Verbal recurrence alone cannot determine the explanatory role of a shared detail.

These texts are hard to read, for scholars and machines alike. Many passages are shorter than a modern paragraph, yet they presuppose an expert reader: verses are cited by a word or two, kabbalistic meaning is signalled by conventional terms (\emph{remez}, \emph{sod}, \emph{derekh ha-emet}) whose weight far exceeds their length, abbreviations compress whole formulae of attribution into two or three letters, and expressions such as ``the Rav'' or ``the way of truth'' carry meaning that the surface text does not state. The passages also mix narrative gloss, halakhic discussion, and kabbalistic commentary in proportions that vary with the verse. This density is what makes the extraction task demanding, and it is why the graphs rely on an ontology designed by scholars of the tradition (Section~\ref{sec:methodology}).

Provenance and licensing. The digitised Ramban text originates from the Chavel edition as re-encoded for the Sefaria open corpus (Creative Commons); the super-commentary transcriptions were prepared by the project team from manuscript facsimiles obtained under research licence from the holding libraries (Escorial, Vatican, London, Parma) and from public-domain printed editions. Bibliographical details for all twenty SCTs, including manuscript identifiers and edition details for printed texts, are given in Appendix~H. All third-party manuscript images are used under the research-only terms of the respective holding libraries; the derived text corpus and the knowledge graphs generated from it are released under Creative Commons Attribution~4.0 for open scholarly reuse.

\section{Related Work}\label{sec:relatedwork}

\subsection{Computational reception and text reuse}\label{subsec:rw-reception}

Computational studies of how texts are received have mostly followed the words. Text-reuse detection finds passages that share wording, from verbatim quotation to paraphrase and allusion, and work on allusive reuse shows how quickly detection degrades as the verbal link weakens \citep{manjavacas2019three,manjavacas2020statistical}. For Rabbinic literature, ACT aligns texts through graph representations of quotations and their contexts \citep{miller2025alignment,miller2026aggadic}, and other work has classified Rabbinic texts by style \citep{tannor2022style} and built language models for Rabbinic Hebrew \citep{shmidman2022berel}. These approaches trace verbal and allusive relationships, but they do not directly represent the conceptual relations through which later commentators may reformulate earlier interpretations. The super-commentaries require a comparison of concepts and their roles, not strings alone.

\subsection{Knowledge graphs from text}\label{subsec:rw-construction}\label{subsec:rw-ontology}

Knowledge-graph construction is classically divided into ontology construction, knowledge extraction, and knowledge fusion \citep{zhong2023comprehensive,ji2021survey} (Supplementary Figure~S1). LLM-based frameworks either keep an explicit schema to guide extraction or let the model produce relations over an open vocabulary \citep{bian2025llm}. Schema-free tools such as KGGen \citep{mo2025kggen} cannot produce graphs comparable across commentators, since the same interpretive relation surfaces under different names; we adopt its two-pass extraction but not its open vocabulary. iText2KG \citep{lairgi2024itext2kg} uses a light schema and merges entities by embedding similarity, which collapses distinct but related concepts, a serious risk where near-synonymous kabbalistic terms must stay apart. In the humanities, \citet{schimmenti2026knowledge} guide LLMs with a fully pre-specified, CIDOC-CRM-compatible ontology and human-validated alignment, which does not scale to twenty super-commentaries, and \citet{pertsas2023humanities} rely on fine-tuned entity recognisers trained on about 10{,}000 labelled sentences per type, which do not exist for medieval Rabbinic Hebrew. For the schema itself, single-pass LLM ontology learning falls short of expert tasks \citep{babaei2023llms4ol}, and fine-tuned approaches need gold standards \citep{lo2024ollm}. We therefore start from a scholar-designed ontology and refine it in context, accepting an addition only if it improves held-out scores (§\ref{subsec:induction}).
A shared ontology makes comparison possible, but it cannot by itself establish whether a relation extracted from an allusive text is warranted. This is why extraction and validation must remain separate steps.

\subsection{Validating graphs with language models}\label{subsec:rw-validation}

Expert annotation of extracted graphs is reliable but does not scale, and structural measures need a gold graph that does not exist here; an LLM judge that checks each relation against its source is the remaining option. The strongest judges are fine-tuned on graph--text pairs \citep{huang2025graphjudge}, which do not exist for this corpus. Zero-shot judges are open to known biases, including self-preference, position, verbosity, and scoring effects, which prompt and scoring design can reduce \citep{li2024llmsjudges,li2025generation}, and on expert tasks they agree with experts less than experts agree with each other \citep{szymanski2025limitations}. We therefore use a judge from a different provider than the generator, compute scores deterministically from its findings, and measure it against domain experts (Section~\ref{sec:experiments}).

\section{From passages to concept graphs}\label{sec:methodology}

\begin{figure*}[!htbp]
    \centering
    \includegraphics[width=0.75\textwidth,alt={Flow diagram of the simpleKG framework. Three inputs at the top, a registry of prompts, the corpus text and a library of ontologies, feed a pipeline of eight numbered stages: punctuation, division into sentence windows, extraction of terms, concepts and relations for each segment, consolidation, merging of concepts with embeddings and a language model, removal of duplicate relations, removal of unconnected concepts, and an optional normalisation of relation names. The pipeline writes the resulting knowledge graphs to a store. Ontology-constrained graphs go to a validator at the lower left, which proposes prompt edits back to the registry; free-format graphs go to an ontology inducer at the lower right, which proposes new concepts and relations back to the ontology library.}]{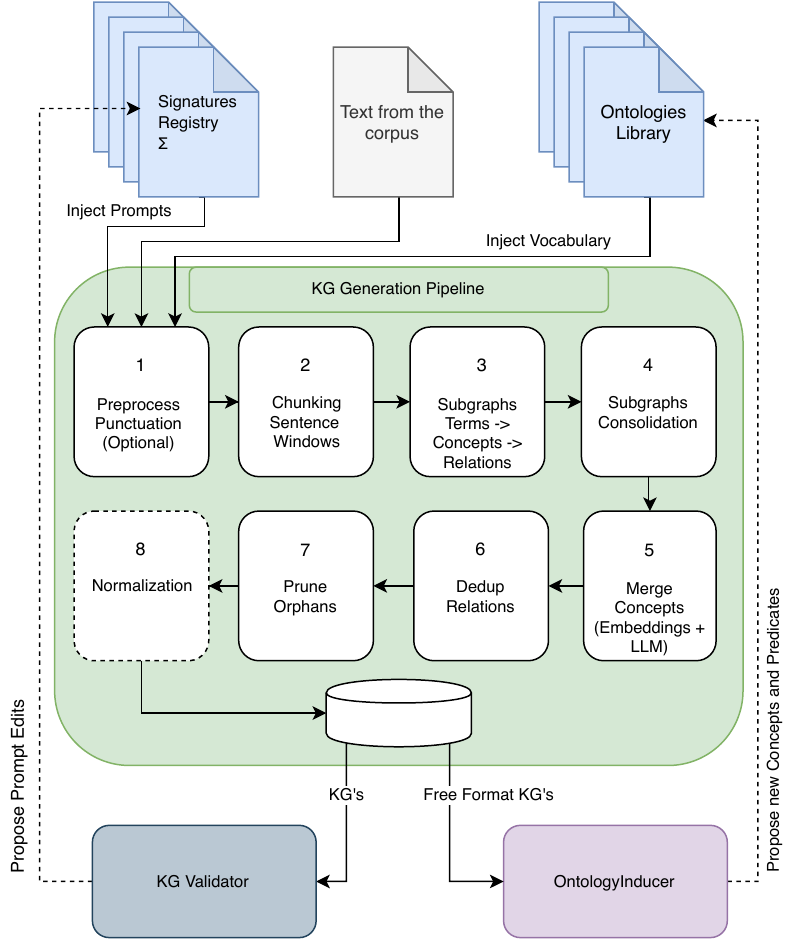}
    \caption{The \texttt{simpleKG} framework. The pipeline (centre) executes eight sequential stages, 1--8, that transform input text into a knowledge graph; stage~8 (dashed) runs only when an ontology is active. The pipeline is fed by a \texttt{Signatures Registry}~$\Sigma$, which injects DSPy prompts, and an \texttt{Ontologies Library}, which injects the typed vocabulary. Both adapters are pluggable per corpus. The resulting KGs are evaluated by two components: the \texttt{KG~Validator} (bottom-left) receives ontology-constrained KGs, and the \texttt{Ontology\allowbreak Inducer} (bottom-right) receives free-format KGs to discover predicates. Their findings feed back to the adapters via the dashed arrows: validator findings propose prompt edits to~$\Sigma$; inducer findings propose new concepts and predicates for the ontology. Pipeline stages are described in §\ref{subsec:pipeline}; the two feedback cycles in §\ref{subsec:induction} and Appendix~K.}
    \label{fig:architecture}
\end{figure*}

Each passage, whether a Ramban comment or a super-commentary response to it, is turned into a knowledge graph by the same pipeline, \texttt{simpleKG} (Figure~\ref{fig:architecture}). The pipeline itself contains nothing specific to the Ramban. What it knows about this corpus comes from two replaceable components: a set of instructions to the language model for each step, and an ontology, the typed vocabulary of concepts and relations that the graphs may use (Appendix~J). A worked example in Appendix~A follows one short Ramban passage from raw text to graph.

\subsection{From text to graph}\label{subsec:pipeline}

The pipeline runs in eight stages (Appendix~J.3). The text is first punctuated, without changing its words, and divided into short overlapping segments, because language models attend unevenly to long inputs \citep{liu2024lostmiddle}. For each segment, the model makes two passes, following \citet{mo2025kggen}: it first lists the concepts the segment mentions, and then, given the segment and that list, states the relations between them. Each concept and relation is stored with the verbatim span of the source that grounds it. The segment graphs are then joined into one graph for the passage: relations that cross segment boundaries are restored, duplicate mentions of the same concept are merged (by embedding similarity, confirmed by the model), repeated relations are removed, and concepts left without any relation are dropped unless the text explicitly defines them. A final stage maps the relation names the model produced onto the canonical names of the ontology. Every intermediate result is kept, so any element of a final graph can be traced back through the stages to the words that produced it.

\subsection{A shared vocabulary}\label{subsec:induction}

Graphs can be compared only if the same interpretive relation receives the same name in every text. The ontology provides that shared vocabulary. Its grain is a trade-off: too coarse, and distinct relations collapse into a generic one, erasing the distinctions the comparison needs; too fine, and it fills with relations that occur in a single text and support no comparison at all.

We start from an ontology designed by scholars of the tradition, \texttt{Ramban\allowbreak Ontology}~(v1), with 20 entity types (for example \texttt{KabbalisticConcept}, \texttt{BiblicalQuotation}, \texttt{RabbinicAuthority}) and 21 relations (for example \texttt{correspondsTo}, \texttt{isCalled}, \texttt{cites}); Appendix~G gives the full specification. Because no scholar can anticipate every relation the texts express, the framework can refine this ontology against the evidence: it collects the relation names the model produces when left unconstrained, proposes additions, and keeps an addition only if it improves the validator's score on a separate set of passages. Additions never overwrite the scholars' decisions, and a round that fails to improve is reverted (Appendix~J.4; full procedure in Appendix~C). For the generator used on the full corpus, the refinement proposed no addition that passed this test, so the full-corpus graphs use the scholar-designed ontology unchanged (§\ref{subsec:exp-multimodel}).

\subsection{What a shared concept measures}\label{subsec:shared-concept}

The comparisons in Section~\ref{sec:results} rest mainly on the concept layer of the graphs. Two concepts count as shared when their normalised names are close in embedding space (cosine similarity of at least 0.80 with \texttt{nomic-embed-text-v2-moe}), so an idea can match across texts even when it is worded differently. A shared concept therefore indicates that two texts engage comparable ideas, even if they use different words. It does not establish shared terminology, the same function in an argument, dependence, or agreement: two commentators can name the same \emph{sefirah} and relate it to different things. Measures built on shared concepts are accordingly read as evidence of common conceptual vocabulary, and every claim beyond that returns to the source spans.

\subsection{Language models used}\label{subsec:llm-use}

Language models are part of the method, not only of its evaluation. Four generators were compared: GPT-4o, Claude Opus~4.6, Qwen~3.5~397B, and gpt-oss-120b. The full corpus was generated with Qwen~3.5~397B. Graphs were scored by Gemini~2.5~Pro, with Claude Opus~4.6 as a second, independent judge, and concepts were embedded with \texttt{nomic-embed-text-v2-moe}. All generation ran at temperature~0. The runs took place between May and July 2026 through the providers' application programming interfaces. Configuration details are given in Appendix~L.

\section{How far can the graphs be trusted?}\label{sec:experiments}

The findings in Section~\ref{sec:results} rest on graphs that a language model produced and another language model scored. Before reading them as evidence about the tradition, we ask how far that is justified. Table~\ref{tab:trust-summary} summarises the evidence. The validator's design is specified in Appendix~K, and the shared configuration and detailed results in Appendix~L.

\begin{table*}[!htbp]
\centering
\small
\caption{Evidence for trusting the graphs. $F_1$ is computed from the validator's findings on 12 held-out passages (precision: share of extracted relations grounded in the text; recall: extracted relations as a share of these plus the relations the judge reports as missing). Detailed results are in the supplement.}\label{tab:trust-summary}
\begin{tabular}{>{\raggedright\arraybackslash}p{0.2\textwidth}>{\raggedright\arraybackslash}p{0.56\textwidth}>{\raggedright\arraybackslash}p{0.14\textwidth}}
\toprule
\textbf{Question} & \textbf{Result} & \textbf{Detail} \\
\midrule
Do experts agree with each other? & Two Ramban scholars, 147 held-out relations: low agreement ($\kappa = 0.18$ on whether a relation is correct) & Appendix~B \\
Does the judge agree with the experts? & On the 60 relations where both experts agree: Claude Opus~4.6 $\kappa = 0.35$, Gemini~2.5~Pro $\kappa = 0.04$; structural accuracy 0.60--0.85 depending on the expert taken as reference & Appendix~B \\
Do two independent judges agree? & Gemini~2.5~Pro and Claude Opus~4.6 on 473 relations: PSA $= 0.42$, $\kappa = 0.36$, PABAK $= 0.81$; most disagreements concern a relation's label, not its existence & Appendix~B \\
How good are the graphs? & Held-out $F_1$ from 0.74 (GPT-4o) to 0.92 (Claude Opus~4.6); 0.88 for Qwen~3.5~397B, the full-corpus generator (95\,\% CI 0.85--0.92, cluster bootstrap over passages) & Table~S4 \\
Does refining the ontology help? & With GPT-4o, $+0.08$ $F_1$ and all relation names mapped to the ontology (from 83\,\%); a second, harmful round was reverted automatically & Table~S3 \\
Is a simpler tool as good? & Whole-passage $F_1$ 0.83, against 0.80 (iText2KG) and 0.80 (KGGen); the confidence intervals overlap & Table~S5 \\
Are the runs reproducible? & Temperature 0; repeated runs give identical graphs & Appendix~L.5 \\
\bottomrule
\end{tabular}
\end{table*}

\subsection{The judge and the experts}\label{subsec:exp-validator}

The first result in the table is a caution against treating the evaluation as a direct measure of historical truth. Two scholars of the tradition, judging the same relations independently, agree only weakly on which are correct. This is not a failure of the annotation but a property of the material: whether a super-commentator ``identifies'' an attribute with a \emph{sefirah} or merely ``associates'' them is itself a matter of interpretation. This disagreement does not establish a fixed ceiling on interpretive accuracy, but it exposes ambiguities for which neither expert can automatically be treated as an unquestioned reference. LLM judges on expert tasks also face documented limits \citep{szymanski2025limitations}. Against the relations on which both experts agree, Claude Opus~4.6 tracks them moderately and Gemini~2.5~Pro, the judge used for all reported scores, hardly at all. A second judge from another provider catches errors at a similar overall rate, but often disagrees about which relations to flag.

We therefore use the judge's scores as a \emph{relative} instrument, not an absolute one. The same judge and instructions make comparisons procedurally consistent, but do not guarantee that errors affect every generator in the same way. Differences between generators or ontologies are therefore indicative under this evaluation protocol, rather than unbiased estimates of historical accuracy. One source of bias was removed before any reported run: in the software framework used, the generator's model could silently be reused as its own judge, and the validator now enforces its own model on every call (Appendix~K).

\subsection{How well the graphs capture the passages}\label{subsec:exp-multimodel}\label{subsec:exp-refinement}\label{subsec:exp-baselines}

Within these limits, the graphs provide a basis for exploratory comparison, especially of shared concepts. On 12 held-out passages, used neither to refine the ontology nor to decide when refinement stops (Appendix~C), four generators reach $F_1$ between 0.74 and 0.92 under the same judge. Qwen~3.5~397B, the full-corpus generator and the stronger of the two open-weight models, reaches 0.88, between the two commercial models. Refining the scholars' ontology helped the weakest generator most (GPT-4o, $+0.08$) and Claude Opus~4.6 only marginally ($+0.02$); for the two open-weight models the proposed additions lowered the score and were reverted (Tables~S3 and S4). Against two general-purpose tools, the framework scores slightly higher, though not significantly, and produces graphs that are both richer than KGGen's and less redundant than iText2KG's (Table~S5). Its main advantages, typed relations and source spans for every element, are not measured by $F_1$ at all.

\subsection{Where the errors lie}\label{subsec:exp-errors}

The errors are not spread evenly. Of the relations one judge flagged and the other did not, 61\,\% differ in the relation's label, 27\,\% in whether the text supports it, and only 12\,\% are outright inventions. The experts found the same pattern from the other side: interpretive relations are wrong far more often than generic ones (\texttt{symbolizes} 44\,\%, \texttt{commentedOn} 40\,\%, \texttt{correspondsTo} 33\,\%, against 12\,\% for \texttt{associatedWith}). The models rarely invent a connection; they more often name it differently from a scholar, and most of all where the naming is itself interpretive.
Aggregate scores therefore conceal an important distinction between recovering a relation and assigning it a historically defensible type.

\subsection{What the graphs can and cannot support}\label{subsec:exp-scope}

These results set the terms on which the findings are read. Comparisons of shared concepts, which most of Section~\ref{sec:results} uses, are less exposed to these errors, since they do not depend on the relation labels where the errors concentrate; this does not show that concept extraction itself is more accurate, and it was not evaluated separately. Claims that rest on particular interpretive relations, such as the frequency of \texttt{correspondsTo} in a work's profile, are weaker and are treated as leads. Scores are compared, not read as accuracies. And no score establishes a historical claim: the graphs indicate where to look, and the reading of the sources decides what is there.
Our analysis accordingly moves from measured similarities to source-grounded interpretations. In particular, concept identity, terminology in the original passage, and the direction and type of the relation must be examined separately.

\section{Findings}\label{sec:results}\label{subsec:res-fullcorpus}\label{subsec:res-grid}

The full run yields non-empty graphs for 383 of the 385 passage-texts (19{,}089 concepts and 25{,}675 relations), arranged as the $38\times21$ grid of Section~\ref{sec:data}. The grid can be read in two directions. Fixing a work and reading across the passages it addresses shows the recurrent features of its reading, where there are any (Section~\ref{subsec:res-commentators}). Fixing a Ramban passage and reading across its responses shows whether the later works stay anchored to him or share concepts his text does not make explicit (Section~\ref{subsec:res-patterns}). Comparing the works as wholes shows which of them belong together (Section~\ref{subsec:res-twins}). In each case the graphs select the passages, and the reading of the texts decides what the selection means. We mark the grade of each claim: counts are measurements, similarities are model-based evidence, and historical conclusions are interpretations that rest on the quoted texts.

\subsection{Each super-commentary reads the Ramban in its own way}\label{subsec:res-commentators}

For each work we pool the graphs of all its responses into a profile and compare it with the Ramban's (Table~\ref{tab:commentator-profiles}). These pooled work-level profiles summarise the concepts found across a work's responses; they are not themselves interpretations of any single passage. Alignment describes conceptual similarity, whereas the proposed interpretive signatures are hypotheses to be checked against the texts. The scores do not establish what is distinctive in a work's historical or doctrinal position.

\begin{table*}[!htbp]
\centering
\small
\caption{Ramban-engagement profiles for the twelve works represented in at least ten Ramban passages, with the Ramban as anchor. \emph{Passages} counts the Ramban passages in which the work appears. \emph{Alignment} is pooled-signature similarity to the Ramban in $[0,1]$, to be read comparatively. \emph{Method marker} records the most frequent explicit hermeneutic term where one dominates, or the most frequent structural relation where explicit terminology is sparse. \emph{Interpretive profile} is a provisional scholarly characterisation of recurring features in the available texts, not a statistically established or necessarily unique signature. Works represented by more than one witness may share a profile; where the available evidence does not establish a distinctive pattern, this is stated explicitly.}\label{tab:commentator-profiles}
\resizebox{\textwidth}{!}{%
\begin{tabular}{lccll}
\toprule
Work & Passages & Alignment & Dominant method marker & Interpretive profile \\
\midrule
Ibn Shuaib (\emph{Be'ur Sodot ha-Ramban}) & 37 & 0.43           & \emph{remez} (allusion)                        & Explicit use of Bahir to explain allusions \\
Keter Shem Tov                     & 37 & 0.24           & \emph{remez} (allusion)                        & Staged, reader-directed exposition of secrets \\
Me'irat Einayim                    & 35 & 0.30           & \emph{remez} (allusion)                        & Explicit terminology: Mercy and Tif'eret \\
Rabbeinu Bahya                     & 32 & 0.26           & \emph{derekh ha-kabbalah}                      & Sefirotic processes: efflux and reception \\
Escorial anonymous                 & 30 & 0.39           & \emph{derekh ha-peshat} (plain sense)          & Patriarchal correspondences: mercy and judgement \\
Recanati (Zikaron Aharon ed.)      & 29 & 0.34           & \emph{remez} (allusion)                        & Juxtaposes alternative kabbalistic readings \\
\emph{Keneged Asara Ma'amarot} Montefiore 487 & 27 & 0.33    & relation: \texttt{correspondsTo}              & Direct, gloss-like decoding of allusions \\
Da'at Chacham, Vatican 107          & 26 & 0.21           & \emph{remez} (allusion)                        & Sequential glosses and paraphrases (same work) \\
Da'at Chacham, Vatican 214          & 26 & 0.22           & \emph{remez} (allusion)                        & Sequential glosses and paraphrases (same work) \\
Keter Shem Tov, London MS           & 17 & 0.28           & \emph{remez} (allusion)                        & Staged exposition (same work) \\
Marginal notes on Ibn Shuaib, Parma 68 MS & 14 & 0.16           & relation: \texttt{correspondsTo}              & Marginal commentary as independent interpretive layer \\
Saporta, Oxford 1645 MS             & 10 & 0.28           & relation: \texttt{correspondsTo}              & No distinctive signature established \\
\midrule
Ramban (anchor)                    & 38 & 1.00 & thin, balanced spread                            & (source) \\
\bottomrule
\end{tabular}%
}
\end{table*}

Two works show part of the range. Ibn Shuaib has the highest pooled alignment to the Ramban among the works in this table (0.43), while explicit references to \emph{Sefer ha-Bahir} recur in his responses. The texts show what kind of addition this is. On the creation of man (RP~1) Ibn Shuaib explains the Ramban's account of the souls ``as written in the \emph{Sefer ha-Bahir}, and the Rav mentioned it here''; on the curses of Leviticus 26 (RP~28) he quotes the Bahir's parable and adds: ``the Rav concealed the matter and did not bring the midrash''. Ibn Shuaib does not move away from the Ramban; he restores the source he believes the Ramban concealed. These examples show the explicit use of an earlier authority to elucidate allusions, but do not establish that Ibn Shuaib relied on this source more than other commentators in every sense.

\emph{Me'irat Einayim} (0.30) adds more, and in a different way. Its recurrent move is a systematic mapping of the Ramban's terms onto the sefirot, including repeated explicit identifications of Tif'eret with mercy. This correspondence is a widely shared kabbalistic convention; the commentator's terminology, rather than doctrinal originality, is the relevant observation. On the shofar (RP~26) it writes: ``the shofar hints at Tif'eret, for it is the attribute of mercy like it; the \emph{teru'ah} hints at the Atarah, which is the attribute of mild judgement''. The work also reports the readings of others and judges them against the Ramban. On the Tower of Babel (RP~6) it quotes a numerological reading received from R. Menahem Ashkenazi and rejects it: ``although these words are good and right for one who understands, the Rav did not intend this, for he did not pursue explaining the secrets of the Torah by way of gematria''. Its elaboration is controlled by an explicit question about what the Ramban meant.
The profile of \emph{Keter Shem Tov} likewise reflects a mode of exposition rather than an isolated pairing of entities: across passages it addresses the reader directly, promises further disclosure, and refers to explanations already provided (e.g. RP~22, 24 and 26). The phrase ``the Rav of blessed memory'' refers to the Ramban and is not itself a designation of Tif'eret. Rabbeinu Bahya, by comparison, sometimes describes the transmission and reception of divine efflux, as in the explanation of incense and the attribute of judgement (RP~23); this is a recurrent type of explanatory movement, not a claim that all divine efflux flows to judgement. In the Escorial anonymous commentary, Jacob is explicitly identified with mercy, while Isaac is associated with judgement (RP~7.1); the earlier proposed identification of Jacob with judgement reverses this textual evidence.

The table distinguishes evidence for a repeated interpretive practice from a claim that such a practice is exclusive to a commentator. Thus the pairing of Tif'eret with Mercy in \emph{Me'irat Einayim}, or of the Great Name with Tif'eret in \emph{Keneged Asara Ma'amarot}, should not be mistaken for doctrinal innovations. The latter commentary often turns the Ramban's hints into short, direct glosses (e.g. RP~13, 16 and 18), though some units develop longer arguments (RP~6 and 30). \emph{Da'at Chacham} similarly offers successive identifications and paraphrases, and both Vatican witnesses are treated as evidence for the same work rather than as independent authorial methods. These parallels suggest a possible family of explanatory practices; establishing such a family would require comparing the behaviour of graph relations across matched passages, not simply their pooled concept sets.

Recanati provides a contrasting practice of juxtaposing different received interpretations. Idel locates his contribution in selecting, combining, and reinterpreting materials from multiple kabbalistic traditions, and documents extensive, often unattributed use of Ibn Gaon's \emph{Keter Shem Tov} \citep[99--103, 110--112]{idel1998recanati}. The profile here records patterns visible in the aligned passages, not the larger organisation of Recanati's intellectual synthesis. On the meaning of the throne (RP~15), he notes that it may allude either to Mercy or to Judgement; elsewhere (RP~27) he explicitly records disagreements among kabbalists. These cases support a profile of presenting alternative traditions, without implying that he always decides between them. The London witness of \emph{Keter Shem Tov} shares the work-level mode of staged disclosure. Its coded kabbalistic terminology belongs to a documented history of deliberate concealment and later editorial decoding, not to an independent interpretive signature \citep[464--466, 478--479]{yisraeli2023keter}. The treatment of these coded terms in transcription and normalisation therefore matters when comparing witnesses. The Parma 68 marginal notes attributed to one commentator can develop their own readings of themes already discussed by Ibn Shuaib (e.g. RP~6, 22 and 31). Their status as a marginal interpretive genre should be distinguished from the unproven claim that every note addresses the Ramban directly rather than its host commentary. Finally, although Saporta elaborates divine actions and ritual effects in some passages (RP~4, 6, 15 and 26), the available graphs do not establish a recurrent or distinctive work-level signature; the table therefore leaves that profile unassigned.

The two witnesses of Da'at Chacham are among the least aligned with the Ramban (0.21 and 0.22) but almost identical to each other (Section~\ref{subsec:res-twins}). Distance from the Ramban is therefore not distance from the tradition: a work can develop far from his vocabulary while remaining stable as a text.
A closer check of explicit Bahir references adds a cross-work observation. Ibn Shuaib names \emph{Sefer ha-Bahir} in nine of thirty-seven available responses, while Recanati does so in four of twenty-nine. All four Recanati responses with such references (RP~10, 12, 14 and 17) also contain Bahir references in Ibn Shuaib. In RP~12 both writers transmit the same parable, to which the Ramban explicitly directs the reader; in RP~14 both connect Moses' raised hands to the ten sefirot and the Bahir, although the Ramban does not name that source there. RP~12 also shows why such overlap cannot settle interpretive agreement: \emph{Keter Shem Tov} identifies \emph{da'at} with Tif'eret, the work associated with Ibn Shuaib explains the phrase through remembrance and renewed attention, and \emph{Da'at Chacham} identifies \emph{da'at} with Yesod. The contrast concerns the meaning assigned to a shared expression, not the mere presence of a term \citep{yisraeliForthcomingTradition}. This is evidence of overlapping textual and interpretive resources, not proof of direct dependence or a common intermediary tradition. The relatively low overall graph similarity in these passages underscores the value of inspecting source-specific overlaps rather than treating one aggregate similarity score as a historical conclusion.

Some profiles rest partly on the interpretive relations, \texttt{correspondsTo} and \texttt{symbolizes}, that the experts found most often wrong (§\ref{subsec:exp-scope}), so the signature identifications are leads to be checked against the texts, as above, rather than classifications.
Profiles capture work-level tendencies but cannot determine whether different commentators interpret a particular Ramban passage similarly; the next comparison returns to passage-level relations.

\subsection{Convergence and fan-out: what the later works share}\label{subsec:res-patterns}\label{subsec:res-cases}

For each Ramban passage we compare two similarities: how close the later responses are to the Ramban ($C_R$) and how close they are to one another ($C_S$). Their difference, $g = C_R - C_S$, is a diagnostic. When $g$ is positive the responses are each closer to the Ramban than to one another: he is the common anchor and they diverge from him (\emph{fan-out}). When $g$ is negative they are closer to one another than to him: they share something his text does not supply (\emph{convergence}). Thresholds of $\pm0.08$, set before the cases were read, flag strong candidates. Other values of this threshold, or of the 0.80 similarity that defines a shared concept, could flag a different set of candidates; we have not tested this sensitivity. These are relative scores of shared concepts, not measures of doctrinal agreement; the thresholds identify candidates for close reading rather than establishing interpretive convergence. A shared concept may perform a different role in each response, while different concepts may be used to explain the same underlying reading.

\begin{figure*}[!htbp]
\centering
\includegraphics[width=\textwidth,alt={Two network graphs side by side. Left, panel (a), RP 32 (the Shema), the Ramban against Da'at Chacham, Vatican 214: three shared concepts in black, labelled in Hebrew (Israel, the name YHWH, the everlasting name); 40 concepts found only in the Ramban, in blue, sit mainly in the upper half, and 53 found only in Da'at Chacham, in orange, mainly in the lower half; the two sides connect only through the three shared concepts. Right, panel (b), RP 6 (the Tower of Babel), the Ramban against Ibn Shuaib: nine shared concepts in black, labelled in Hebrew (their sin, their father's sin, their father, Sodom, cutting the shoots, the unique Name, the dispersion, measure for measure, the tower), with 18 concepts only in the Ramban in blue and 18 only in Ibn Shuaib in orange; the shared concepts form a connected core at the centre, with both sides attached to it.}]{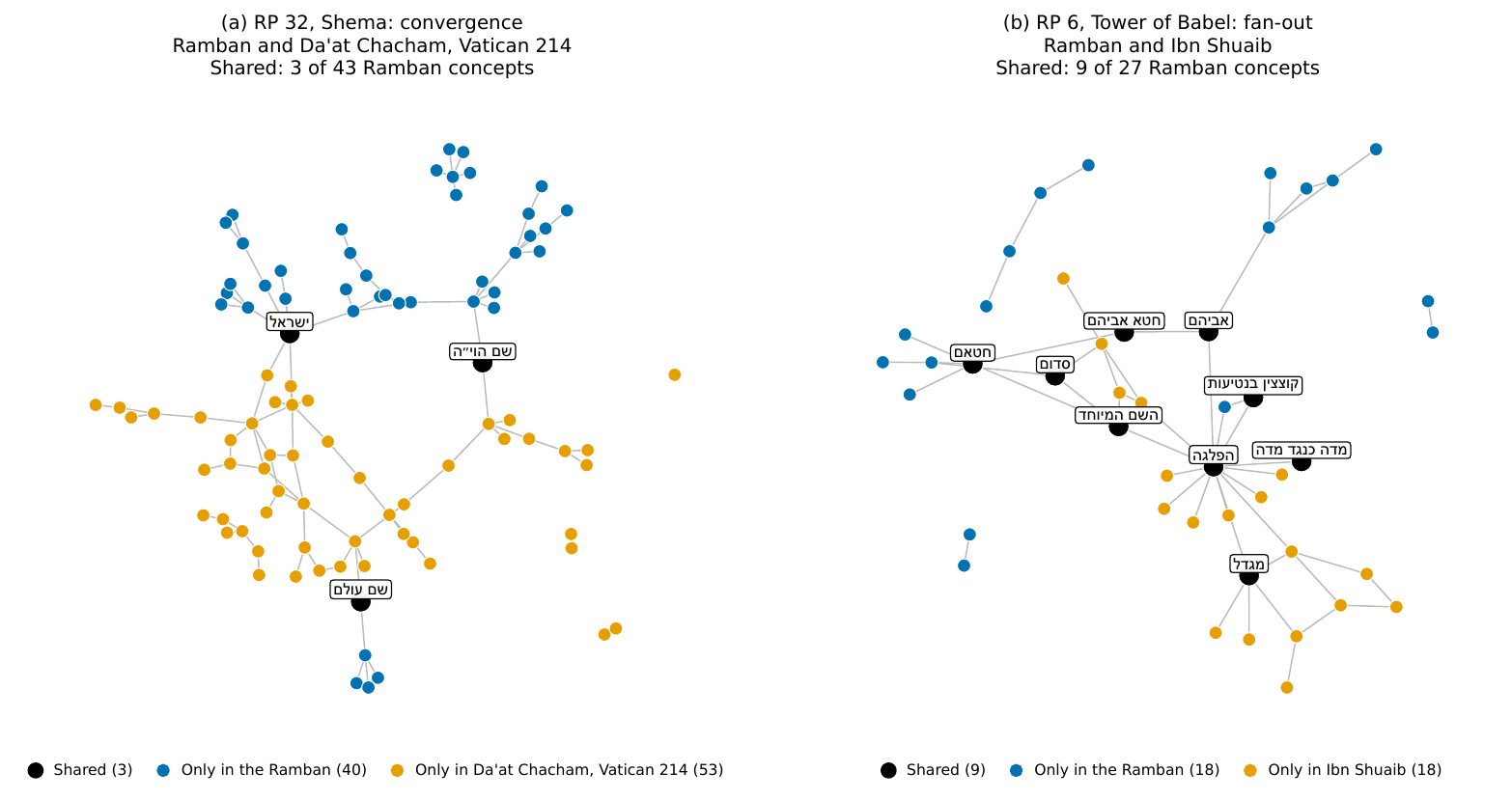}
\caption{Concept-overlap graphs for the two cases read in this section. Each panel compares the Ramban graph on a passage with one response: black nodes are shared concepts, blue nodes occur only in the Ramban, and orange nodes only in the response. \emph{(a) RP~32 (Shema), a convergence passage.} Da'at Chacham, Vatican~214, retains 3 of the Ramban's 43 concepts; the shared core is two divine names and Israel. \emph{(b) RP~6 (Tower of Babel), a fan-out passage.} Even the closest response, Ibn Shuaib, retains 9 of the Ramban's 27 concepts. A single pair illustrates shared and unique material; the passage-level measure $g$ is computed over all responses.}\label{fig:overlap-cases}
\end{figure*}

The gap is positive in 27 of the 36 passages where it is defined, and six of these cross the fan-out threshold. Four passages meet the convergence threshold; two of them are artefacts of the data (in RP~1, for example, the Ramban cell held a comment on a different verse), and two survive inspection: RP~29, on the ``swallowing'' of the sacred vessels (Num 4:20), and RP~32, the Shema (Deut 6:4). The usual pattern is therefore fan-out from the Ramban, and convergence is rare. We read one case of each (Figure~\ref{fig:overlap-cases}).

\paragraph{The Shema (RP~32): convergence by decoding.}
Here the responses are closer to one another than to the Ramban ($C_R = 0.203$, $C_S = 0.298$, $g = -0.095$), and the Ramban is the most peripheral of the eight works. His graph is dominated by grammar and cross-reference; theirs share the unification of the sefirot, Tif'eret, the six extremities and the Atarah (a designation of the lowest sefirah, associated with the Shekhinah). Read alone, the scores suggest that the later works converge on an elaboration that has little to do with him.

The texts show something else. The Ramban's comment begins from a grammatical observation, that the verse says ``the Lord \emph{our} God'' where Deuteronomy usually says ``your God'', and then sets out a chain of hints without decoding any of them: God did great things with Moses ``to make Himself a name of \emph{tif'eret}''; the dalet of \emph{ehad} is written large ``to hint at'' the ``everlasting name'' of Isaiah 63:12; ``therefore'' the sages instituted \emph{barukh shem kevod malkhuto} after the verse, with the parable of the king's daughter who smelled the dish, ``for Moses said it in the Torah by way of hint''. Each response follows this chain step by step and decodes it. Ibn Shuaib reads ``our God'' as a hint at the final heh, \emph{tif'eret} as ``Tif'eret Yisrael, with the six extremities included with it'', and the king's daughter as the dalet, which would be pained to be left out of the unification. Rabbeinu Bahya maps the words of the verse onto the ten sefirot (``the Lord'' the three upper ones, ``our God'' Gedulah and Gevurah, ``the Lord'' Tif'eret, and in ``one'' all ten are included) and makes the large dalet ``the great golden crown''. \emph{Me'irat Einayim} reports a sefirotic mapping attributed to Saporta but also critically examines an aspect of that received interpretation; its use of shared concepts therefore does not establish agreement with its source.

The conceptual convergence, then, does not represent a departure from the Ramban: the later works elaborate the hints he leaves unexplained, although they do not always endorse the same interpretation. His low score reflects his surface vocabulary, not his role in the reception. Two further observations qualify the picture. First, the grammatical point that the graphs treat as left behind is taken up and transformed. Keter Shem Tov explains ``our God'' as Moses including himself in the acceptance of the kingdom of heaven, ``for had he said `your God' he would seem to have excluded himself''; Ibn Shuaib turns the same observation into a kabbalistic sign. Second, the works name the authorities through which they decode. \emph{Me'irat Einayim} presents its mapping as ``the tradition of Saporta''; Da'at Chacham adds ``and in the Zohar I found [a reading] according to the view of the Rav'' and quotes it at length; and Recanati notes that ``the kabbalists disagree whether the order is from above to below or from below to above''. The readings draw on named traditions, whose earlier history we do not trace here; shared terms and attributions can coexist with disagreement over how the Ramban should be understood.

\paragraph{The Tower of Babel (RP~6): fan-out around a shared core.}
Here the pattern is reversed ($C_R = 0.274$, $C_S = 0.186$, $g \approx 0.09$): the thirteen responses are each closer to the Ramban than to one another. The Ramban's comment is again a set of hints: ``one who knows the meaning of `name' will understand their intention'' from ``let us make us a name'', and ``will know how great the measure'' they planned for the tower; their sin was ``cutting the shoots'' (a kabbalistic image of separating divine powers), like their father Adam's; they were punished ``by His great name''; ``and the enlightened will understand''.

At the level of concepts the responses go in different directions. Da'at Chacham, Saporta and Ibn Shuaib develop the ``measure'' through the \emph{Shi'ur Komah}, the ancient tradition of the measurements of the divine body; Keter Shem Tov and \emph{Me'irat Einayim} read ``name'' as the Atarah, the last sefirah; Recanati and the \emph{Likutim} map the city and tower onto the divine structure; Rabbeinu Bahya and the Escorial anonymous follow a midrashic and historical account of the builders' sin. These different vocabularies are what the graphs measure.

The texts also show what the scores do not: under the different vocabularies most responses decode ``cutting the shoots'' in the same way. For Keter Shem Tov the builders wanted ``to make them a name, that is the Atarah, \ldots\ to be conducted by this'' alone; for Ibn Shuaib ``they took the attribute of the Name for themselves, saying that it governs the world and suffices for them''; for Saporta ``the governance of all the matters of this world is through the Atarah, and there is no need for the rest of the structure''; for Bahya they ``made a beginning from the end'', the last attribute, ``and did not believe in what is above it''; for Recanati ``they wanted to separate the Shekhinah from the structure''. In these responses the sin is the isolation of the last sefirah from the rest; whether this reading is older than the super-commentaries lies outside the scope of this study. The responses also share the Ramban's reticence: Keter Shem Tov ``was not permitted to hint at this'', Ibn Shuaib ``cannot explain more'', and Recanati ``cannot interpret''. Fan-out in the measured concepts thus coexists with a shared reading of the central secret. The works differ in their explanatory vocabularies and in how they elaborate received interpretations, including the ``measure'', where the Ramban offered little explicit guidance.

A second candidate for convergence, RP~29 (Num 4:20), concerns the concealment of the sacred vessels. Across the responses, the concealment is read as reflecting the withdrawal or concealment of divine glory, an interpretation already grounded in the Ramban's own formulation. Particularly close explanatory sequences in \emph{Keter Shem Tov} and Recanati, and in the anonymous Escorial and Vatican~214 commentaries, suggest more specific textual or interpretive affinities. These parallels require distinct historical interpretations. The close pairing of \emph{Keter Shem Tov} and Recanati should be read against their documented relationship: Recanati drew extensively on Ibn Gaon, sometimes without attribution \citep[110--112]{idel1998recanati}. The particular passages still require textual comparison before any direction or mechanism of borrowing is claimed; other pairings may reflect shared sources or explanatory conventions.

The three cases clarify the relationship between conceptual scores and interpretation. In RP~32, shared vocabulary coexists with disagreement over the decoding of the Ramban's hints; in RP~6, divergent vocabularies coexist with a broadly shared reading of the central secret. In RP~29, widespread common content must be distinguished from the closer textual affinities of particular pairs. The graphs identify contrasting patterns across the corpus, while close reading determines whether these patterns involve preservation, elaboration, critical engagement or possible textual transmission.
This distinction also depends on what counts as an independent textual unit, a question taken up by the manuscript comparisons below. %

\subsection{The graphs recover the witnesses of one work}\label{subsec:res-tradition}\label{subsec:res-twins}

\begin{figure}[!htbp]
\centering
\includegraphics[width=0.85\linewidth,alt={Horizontal dendrogram of sixteen texts, the Ramban and fifteen super-commentary texts, with distance from 0 to about 0.9 on the horizontal axis. The closest joins are Keter Shem Tov with its London manuscript (about 0.25), the Vatican 214 anonymous unit with the Escorial anonymous (about 0.27), and Da'at Chacham, Vatican 214, with Da'at Chacham, Vatican 107 (about 0.36). Ibn Shuaib joins the Likutim at about 0.56, and Saporta joins the two anonymous units at about 0.58. The Ramban joins the marginal notes on Da'at Chacham, Vatican 214, at about 0.64. Recanati joins the Keter Shem Tov pair at about 0.67, and Keneged Asara Ma'amarot joins the Da'at Chacham pair at about 0.72. Rabbeinu Bahya joins at about 0.78, and the marginal notes on Ibn Shuaib, Parma 68, join last, at about 0.87.}]{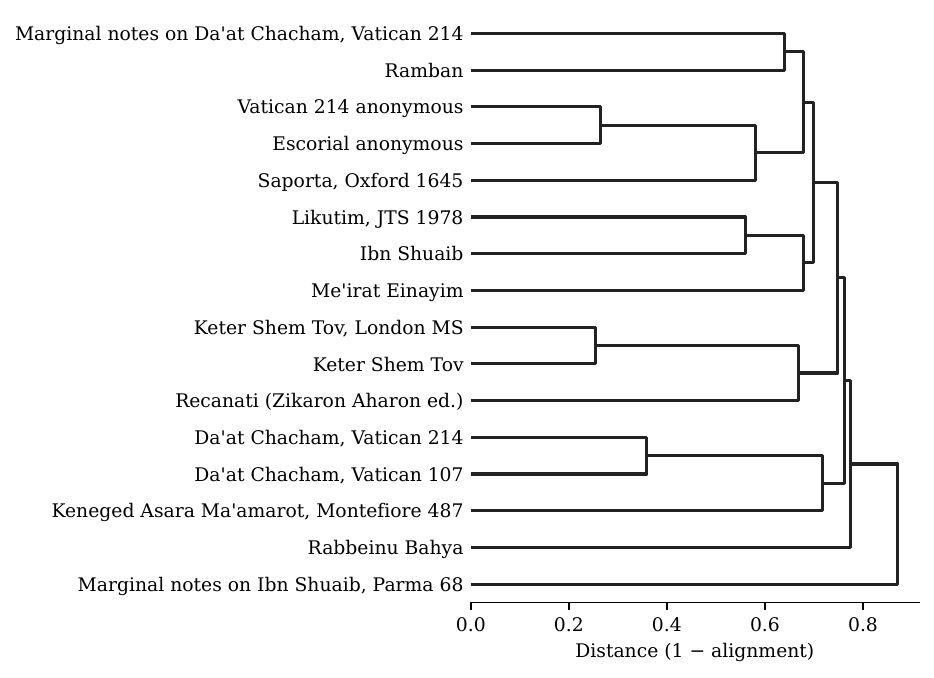}
\caption{Hierarchical clustering of the fifteen super-commentary texts represented in more than one passage, plus the Ramban, on pooled concept embeddings. Pairwise distance is one minus symmetric alignment at $\theta=0.80$, with average-linkage clustering; branch height, not leaf order, encodes dissimilarity. This distance is not the signature similarity reported in the text, and the two measures can rank close neighbours differently: here the Vatican~214 anonymous unit joins the Escorial anonymous, whereas by signature similarity it is closest to the marginal notes on \emph{Da'at Chacham}. The figure identifies neighbourhoods and candidate kinships; historical interpretation requires the texts.}\label{fig:tradition-map}
\end{figure}

Comparing whole works asks a different question: which of them belong together? We cluster the fifteen works represented in more than one passage, together with the Ramban, on their pooled concept embeddings (Figure~\ref{fig:tradition-map}). The result is a graded field of neighbourhoods, not a family tree, and branch height, not left-to-right order, carries the distances.

The closest pairs are known witnesses of one work. Keter Shem Tov and its London manuscript have similarity 0.939, against a corpus mean of 0.81, and the two witnesses of Da'at Chacham, Vatican 107 and Vatican 214, have 0.984. No information about witnesses enters the clustering, so recovering these pairs is a check that the graphs preserve the identity of a text across its copies.

The corpus also contains marginal notes on \emph{Da'at Chacham} in Vatican~214, distinct from the two witnesses of the main work. An incorrect manuscript number in the initial catalogue data was identified and corrected through philological inspection by a member of the research team, independently of the graph analysis. The notes are retained as a separate textual unit for comparison.

The marginal notes raise two further questions that the graphs can pose but not answer. First, an anonymous unit assigned in the project transcription to Vatican 214 (the current folio attribution, ff.~199v--213v, requires verification against the manuscript and the anonymous commentary discussed by \citet{idel1979unknown}) has signature similarity 0.94 with the notes, and the two never respond to the same Ramban passage: the notes appear in eight passages and the anonymous unit in seven others. This complementary distribution and conceptual similarity invite investigation but establish neither a single work nor a direction of textual transmission. Moreover, Yisraeli's account of the later separation of \emph{Da'at Chacham} from combined witnesses of \emph{Keter Shem Tov} shows that independent presentation in a codex cannot by itself establish independent composition \citep[478--479]{yisraeli2023keter}. The Vatican units remain separate analytical units pending verification of their locations and manuscript collation. Second, in the clustering the Ramban's nearest neighbour is the marginal notes, not any of the commentaries. The notes frequently follow his wording and explanatory sequence closely, notably in RP~21--23. This offers a possible explanation of the proximity, although the clustering cannot establish its cause.

Together, the results show how manuscript relationships and interpretive similarities must be distinguished. The graphs recover known pairs of witnesses and identify additional affinities for investigation; documentary correction and any stronger historical claim depend on the texts and manuscripts, not on graph distance alone.

\section{Discussion}\label{sec:discussion}

The paper set out with two aims, one historical and one methodological. We take them in turn, and then state the limits of both.

\subsection{What the graphs add to the study of the Ramban's reception}\label{subsec:disc-reception}

The clearest result is that reception was selective not merely in which concepts were retained, but in how their explanatory roles changed. A grammatical observation could become the basis for a sefirotic interpretation, while another passage might be read through an earlier authoritative source. Historical and philological studies have already reconstructed the transmission of the Ramban's esoteric teachings, disagreements over the authority of inherited interpretations, and the rewriting of those traditions through compilation and editorial change \citep{halbertal2006way,idel1983noKabbalisticTradition,abrams1996orality,yisraeli2023keter,yisraeliForthcomingTradition,lawee2017genre}. Idel's account of Recanati further shows why the selection and arrangement of inherited sources, rather than novelty of individual concepts, can be historically consequential \citep[99--103]{idel1998recanati}. The present contribution is to trace the distribution of these processes across an aligned corpus, distinguish recurrent patterns from exceptional cases, and identify passages for philological examination. Because our principal similarity measure compares shared concepts, it does not by itself reconstruct the intellectual organisation of a work.

The passage-level comparison reveals a corpus-wide distinction that is difficult to establish by reading individual works separately. For most Ramban passages, the later responses stay closer to him than to one another: his text is the common anchor, and the tradition develops from it in several directions. For a few passages, notably the Shema, the responses converge on an emphasis that the Ramban's own comment barely contains. Such rows may reflect a shared later interpretive repertoire, but conceptual convergence by itself cannot distinguish inherited knowledge, shared explanatory practices, the pull of the verse or direct textual transmission. Philological investigation must decide among these possibilities; the graphs show where such investigation may be productive.

At the level of whole works, the substantially represented commentaries combine the Ramban's material with further sources and concepts. Most show recurrent interpretive practices, from Ibn Shuaib's explicit use of the \emph{Sefer ha-Bahir} to the staged exposition of \emph{Keter Shem Tov} and Recanati's juxtaposition of alternative readings. Such practices are not necessarily exclusive to one work, and for some, such as Saporta, no distinctive profile is established. Because these profiles rest partly on the interpretive relations where the graphs are least reliable (§\ref{subsec:exp-scope}), we offer them as characterisations to be tested against the texts, not as classifications. Finally, the comparison recovers witnesses known to transmit one work as near neighbours, and it identifies a pair of units associated with Vatican 214 for further textual and folio verification (§\ref{subsec:res-twins}).

\subsection{What the study shows about the method}\label{subsec:disc-affordances}\label{subsec:disc-bridge}

For this corpus, language-model-generated graphs support exploratory comparison of medieval Rabbinic Hebrew without labelled training data, provided their scores are treated as judge-dependent indicators rather than absolute measures of accuracy. The more general lesson concerns evaluation. Two experts on this tradition agree only weakly about which extracted relations are correct, so there is no single gold standard against which an interpretive extraction could be scored. Evaluating such work means measuring the disagreement among experts and placing the model within it, not reporting accuracy against one annotator.

The errors also fall where the interpretation does. The models rarely invent connections; they more often name a relation differently from a scholar, and most often where naming is itself an interpretive act. This argues for comparing texts at the level of shared concepts, which depends less on these disagreements, and for treating claims built on particular interpretive relations as leads for scholarly review.

Finally, every concept and relation keeps the source span it was drawn from. A scholar can therefore move from a corpus-level pattern back to the passages behind it and make two separate checks: whether the extraction is grounded in the text, and whether close reading supports the historical interpretation proposed for the pattern. The framework organises evidence at corpus scale; the meaning of a convergence, a divergence, or a manuscript affinity remains a matter of philological judgement. The conceptual layer also complements graph-based text-reuse alignment such as ACT \citep{miller2025alignment,miller2026aggadic}; combining the two to detect reuse across authors is future work.

\subsection{Limitations}\label{subsec:disc-limits}

The reliability argument rests on twelve held-out passages and 147 expert-annotated relations, a small sample. The relation labels are especially vulnerable to conflating conceptual equivalence with directed actions such as influence, reception, or the transmission of divine efflux. Concept extraction and merging were not evaluated separately from the relations. Likewise, normalising concepts may obscure historically significant terminological choices even when the conceptual mapping is plausible. The judge behind every reported score is a general-purpose model that agrees poorly with the experts; a judge adapted to the domain \citep{huang2025graphjudge} might improve the evaluation, and the baseline comparison uses the framework's own judge, which may favour it. On the historical side, the 38 Ramban passages were chosen where the tradition is most active, so the grid over-represents interpretive activity, and works outside the twelve-SCT core are thinly attested. A shared concept does not establish dependence, chronology, or agreement: as the competing explanations of \emph{da'at} in RP~12 illustrate, the same word may also refer to distinct objects or functions. Interpretation remains dependent on close reading and, for manuscripts, collation. Finally, the study uses one corpus. The framework is designed so that a new corpus needs only new instructions and a new ontology (Appendix~J), but whether its results hold for other genres, languages, and traditions must be tested corpus by corpus.

\section{Conclusion}\label{sec:conclusion}

This paper asked how the Ramban's commentary was received by the kabbalists who wrote on it, and whether language models can help answer that question at the scale of a whole tradition. Concept graphs of 385 aligned passages, checked against the judgement of experts, show a reception that was selective rather than uniform. The later works developed his kabbalistic hints, sometimes transforming even grammatical observations into elements of esoteric interpretation. Most often they used his text as a common anchor and developed it in different directions; for a few passages their shared concepts point to interpretations that require closer attention to the earlier sources they invoke. Their reception can involve both the selection of material and the transformation of its interpretive role. These comparisons neither measure the complete organisation of each commentator's knowledge nor establish historical transmission or decide how individual passages should be read. They locate patterns of selection, conceptual affinity and difference across the corpus, enabling closer study of possible shared interpretive traditions while keeping the interpretation of textual evidence with the reader of the sources.

\paragraph{Supplementary material}
The supplementary material prepared for this article (Appendices A--L) gives a worked example of the pipeline, the statistics behind the trust section, the ontology and its induction, the work-level profiles and worked readings in full, cost and latency, the bibliographical details of the super-commentaries, and the design of the framework, the validator and the experiments.

\paragraph{Acknowledgements}
\ifanon Acknowledgements of individuals are withheld for anonymous review.\else The authors thank Tsvi Kuflik for his careful reading of successive drafts and his constructive comments throughout the writing of this article.\fi

The use of language models as part of the method is described in Section~\ref{subsec:llm-use}. In addition, Anthropic's Claude (Claude Opus~4.8, Claude Fable~5 and Claude Opus~5.5, used through Claude Code between May and November 2026) assisted with the writing. It summarised project documentation, drafted and revised prose, drafted first versions of the close readings and the English translations in Section~\ref{sec:results} from the corpus texts, organised the work plans, and conducted an internal review of the manuscript. The close readings and translations were checked and corrected by the authors, and all content, including factual claims, citations, numbers and interpretations, was reviewed and verified by the authors, who take responsibility for it. No AI tool designed the Ramban ontology or produced the experimental results.

\paragraph{Data availability statement}
The digitised text corpus, the aligned super-commentary passages, the machine-readable ontology and the knowledge graphs underlying this article are openly available on Zenodo at \ifanon [DOI withheld for anonymous review]\else\url{https://doi.org/10.5281/zenodo.22539818}\fi, under a Creative Commons Attribution~4.0 licence. \ifanon The code of the \texttt{simpleKG} framework is available at [link withheld for anonymous review].\else The code of the \texttt{simpleKG} framework will be made publicly available on publication.\fi{} Manuscript images are not included, in accordance with the research-only terms of the holding libraries.

\paragraph{Ethical standard}
This research analyses medieval texts and did not involve human participants. The expert judgements reported in Section~\ref{sec:experiments} were made by scholars of the tradition as part of the research and record no personal data.

\paragraph{Author contributions}
\ifanon Withheld for anonymous review.\else Conceptualisation: H.M., E.B. Methodology: H.M., E.B. Software: H.M. Formal analysis: H.M. Investigation: H.M., E.B., M.L. Data curation: E.B. Validation: O.Y. Writing -- original draft: H.M. Writing -- review and editing: H.M., E.B., O.Y., M.L.\fi

\paragraph{Funding statement}
This research received no specific grant from any funding agency, commercial or not-for-profit sectors.

\paragraph{Competing interests}
The authors declare none.

\printbibliography

@article{zhong2023comprehensive,
  author    = {Zhong, Lingfeng and Wu, Jia and Li, Qian and Peng, Hao and Wu, Xindong},
  title     = {A Comprehensive Survey on Automatic Knowledge Graph Construction},
  journal   = {ACM Computing Surveys},
  volume    = {56},
  number    = {4},
  pages     = {Article 94},
  year      = {2023},
  publisher = {Association for Computing Machinery},
}

@article{ji2021survey,
  author    = {Ji, Shaoxiong and Pan, Shirui and Cambria, Erik and Marttinen, Pekka and Yu, Philip S.},
  title     = {A Survey on Knowledge Graphs: Representation, Acquisition, and Applications},
  journal   = {IEEE Transactions on Neural Networks and Learning Systems},
  volume    = {33},
  number    = {2},
  pages     = {494--514},
  year      = {2021},
  publisher = {IEEE},
}

@article{lairgi2024itext2kg,
  author    = {Lairgi, Yassir and Moncla, Ludovic and Cazabet, R{\'e}my and Benabdeslem, Khalid and Cl{\'e}au, Pierre},
  title     = {{iText2KG}: Incremental Knowledge Graphs Construction Using Large Language Models},
  journal   = {arXiv preprint arXiv:2409.03284},
  year      = {2024},
  note      = {Code at https://github.com/AuvaLab/itext2kg},
}

@inproceedings{mo2025kggen,
  author    = {Mo, Belinda and Yu, Kyssen and Kazdan, Joshua and Petrick, Proud and Yu, Charilaos and Krishnan, Aaryan and Vidgen, Bertie and Koyejo, Sanmi and Suzgun, Mirac},
  title     = {{KGGen}: Extracting Knowledge Graphs from Plain Text with Language Models},
  booktitle = {{Advances in Neural Information Processing Systems (NeurIPS) 38}},
  year      = {2025},
  note      = {Code at https://github.com/stair-lab/kg-gen},
}

@inproceedings{bian2025llm,
  author    = {Bian, Yuhao},
  title     = {{LLM}-Empowered Knowledge Graph Construction: A Survey},
  booktitle = {{Proceedings of the 2025 International Conference on Artificial Intelligence and Statistics (ICAIS)}},
  year      = {2025},
}

@inproceedings{babaei2023llms4ol,
  author    = {Babaei Giglou, Hamed and D'Souza, Jennifer and Auer, S{\"o}ren},
  title     = {{LLMs4OL}: Large Language Models for Ontology Learning},
  booktitle = {{The Semantic Web -- ISWC 2023}},
  series    = {Lecture Notes in Computer Science},
  year      = {2023},
  publisher = {Springer},
  note      = {arXiv:2307.16648},
}

@inproceedings{lo2024ollm,
  author    = {Lo, Andy and Jiang, Albert Q. and Li, Wenda and Jamnik, Mateja},
  title     = {End-to-End Ontology Learning with Large Language Models},
  booktitle = {{Advances in Neural Information Processing Systems (NeurIPS) 37}},
  year      = {2024},
  note      = {Code at https://github.com/andylolu2/ollm},
}

@inproceedings{huang2025graphjudge,
  author    = {Huang, Haoyu and Chen, Chen and Sheng, Conghui and Li, Zhongjiang and Zhang, Yangqiu},
  title     = {Can {LLMs} be Good Graph Judge for Knowledge Graph Construction?},
  booktitle = {{Proceedings of the 2025 Conference on Empirical Methods in Natural Language Processing (EMNLP)}},
  pages     = {10929--10948},
  year      = {2025},
  publisher = {Association for Computational Linguistics},
  note      = {Code at https://github.com/hhy-huang/GraphJudge},
}

@inproceedings{li2025generation,
  author    = {Li, Dawei and Jiang, Bohan and Huang, Liangjie and Beigi, Alimohammad and Zhao, Chengshuai and Tan, Zhen and Bhattacharjee, Amrita and Jiang, Yuxuan and Chen, Canyu and Wu, Tianhao and Shu, Kai and Cheng, Lu and Liu, Huan},
  title     = {From Generation to Judgment: Opportunities and Challenges of {LLM}-as-a-Judge},
  booktitle = {{Proceedings of the 2025 Conference on Empirical Methods in Natural Language Processing (EMNLP)}},
  pages     = {2757--2791},
  year      = {2025},
  publisher = {Association for Computational Linguistics},
}

@misc{li2024llmsjudges,
  author    = {Li, Haitao and Dong, Qian and Chen, Junjie and Su, Huixue and Zhou, Yujia and Ai, Qingyao and Ye, Ziyi and Liu, Yiqun},
  title     = {{LLMs}-as-Judges: A Comprehensive Survey on {LLM}-based Evaluation Methods},
  howpublished = {arXiv preprint arXiv:2412.05579},
  year      = {2024},
  note      = {Repository at https://github.com/CSHaitao/Awesome-LLMs-as-Judges},
}

@inproceedings{szymanski2025limitations,
  author    = {Szymanski, Annalisa and Ziems, Noah and Eicher-Miller, Heather and Li, Tony Toh-Jen and Jiang, Meng and Metoyer, Ronald A.},
  title     = {Limitations of the {LLM}-as-a-Judge Approach for Evaluating {LLM} Outputs in Expert Knowledge Tasks},
  booktitle = {{Proceedings of the 30th International Conference on Intelligent User Interfaces (IUI '25)}},
  year      = {2025},
  publisher = {Association for Computing Machinery},
}

@inproceedings{pertsas2023humanities,
  author    = {Pertsas, Vayianos and Leontaridis, Panagiotis and Kasapaki, Marialena and Constantopoulos, Panos},
  title     = {A Knowledge Graph for Humanities Research},
  booktitle = {{Digital Humanities 2023 -- Book of Abstracts}},
  year      = {2023},
}

@inproceedings{shmidman2022berel,
  author    = {Shmidman, Avi and Guedalia, Joshua and Shmidman, Shaltiel and Shmidman, Cheyn Shmuel and Handel, Eli and Koppel, Moshe},
  title     = {Introducing {BEREL}: {BERT} Embeddings for {Rabbinic}-Encoded Language},
  booktitle = {{Proceedings of the 2022 DICTA Workshop on Rabbinic-Hebrew NLP}},
  year      = {2022},
  note      = {arXiv:2208.01875},
}

@article{tannor2022style,
  author    = {Tannor, Shlomo and Dershowitz, Nachum and Lavee, Moshe},
  title     = {Style Classification of {Rabbinic} Literature for Detection of Lost {Midrash Tan\d{h}uma} Material},
  journal   = {Journal of Data Mining and Digital Humanities},
  year      = {2023},
  note      = {arXiv:2211.09710},
}

@inproceedings{manjavacas2019three,
  author    = {Manjavacas, Enrique and Long, Brian and Kestemont, Mike},
  title     = {On the Feasibility of Automated Detection of Allusive Text Reuse},
  booktitle = {{Proceedings of the 3rd Joint SIGHUM Workshop on Computational Linguistics for Cultural Heritage, Social Sciences, Humanities and Literature (LaTeCH-CLfL)}},
  year      = {2019},
  publisher = {Association for Computational Linguistics},
}

@inproceedings{manjavacas2020statistical,
  author    = {Manjavacas, Enrique and Karsdorp, Folgert and Kestemont, Mike},
  title     = {A Statistical Foray into Contextual Aspects of Intertextuality},
  booktitle = {{Proceedings of the Workshop on Computational Humanities Research (CHR)}},
  pages     = {77--98},
  year      = {2020},
}

@article{miller2025alignment,
  author    = {Miller, Hadar and Kuflik, Tsvi and Lavee, Moshe},
  title     = {Text Alignment in the Service of Text Reuse Detection},
  journal   = {Applied Sciences},
  volume    = {15},
  number    = {6},
  pages     = {3395},
  year      = {2025},
  publisher = {MDPI},
  doi       = {10.3390/app15063395},
}

@article{miller2026aggadic,
  author    = {Miller, Hadar and Kuflik, Tsvi and Lavee, Moshe},
  title     = {Automatic Detection of Complex Quotation Patterns in {A}ggadic Literature},
  journal   = {Cogent Arts \& Humanities},
  volume    = {13},
  number    = {1},
  pages     = {2657191},
  year      = {2026},
  publisher = {Taylor \& Francis},
  doi       = {10.1080/23311983.2026.2657191},
}

@phdthesis{miller2026phd,
  author  = {Miller, Hadar},
  title   = {Automatic Text Reuse Detection Tailored to the {Aggadic} Literature},
  school  = {University of Haifa, Department of Information Systems},
  year    = {2026},
  address = {Haifa, Israel},
  type    = {Ph.D. dissertation, Monograph},
  note    = {Supervisors: Tsvi Kuflik, Moshe Lavee.},
}

@article{schimmenti2026knowledge,
  author    = {Schimmenti, Andrea and Pasqual, Valentina and Vitali, Fabio and Van Erp, Marieke},
  title     = {Knowledge graphs generation from cultural heritage texts: combining {LLMs} and ontological engineering for scholarly debates},
  journal   = {Journal of Documentation},
  volume    = {82},
  number    = {7},
  pages     = {206--250},
  year      = {2026},
  publisher = {Emerald Publishing Limited},
}

@book{yisraeli2020nachmanides,
  author    = {Yisraeli, Oded},
  title     = {{R. Moses b. Nachman (Nachmanides)}: Intellectual Biography},
  address   = {Jerusalem},
  publisher = {Magnes Press},
  year      = {2020},
}

@incollection{idel1998nahmanides,
  author    = {Idel, Moshe},
  title     = {{Nahmanides}: Kabbalah, Halakhah, and Spiritual Leadership},
  booktitle = {{Jewish} Mystical Leaders and Leadership in the Thirteenth Century},
  editor    = {Idel, Moshe and Ostow, Mortimer},
  pages     = {15--96},
  address   = {Northvale, NJ},
  publisher = {Jason Aronson},
  year      = {1998},
}

@book{halbertal2006way,
  author    = {Halbertal, Moshe},
  title     = {By Way of Truth: {Nahmanides} and the Creation of Tradition},
  address   = {Jerusalem},
  publisher = {Shalom Hartman Institute},
  year      = {2006},
  note      = {Hebrew},
}

@incollection{idel1983noKabbalisticTradition,
  author    = {Idel, Moshe},
  title     = {`{We} Have No Kabbalistic Tradition on This,' {Rabbi Moses Nahmanides (Ramban)}: Explorations in His Religious and Literary Virtuosity},
  booktitle = {{Rabbi Moses Nahmanides (Ramban)}: Explorations in His Religious and Literary Virtuosity},
  editor    = {Twersky, Isadore},
  pages     = {51--73},
  address   = {Cambridge, MA},
  publisher = {Harvard University Press},
  year      = {1983},
}

@article{yisraeli2023keter,
  author  = {Yisraeli, Oded},
  title   = {The History of {Keter Shem Tov} by {R. Shem Tov Ibn Gaon}: A Chapter in the Reception of {Nahmanides'} {Kabbalah} in the Thirteenth and Fourteenth Centuries},
  journal = {Tarbiz},
  volume  = {89},
  number  = {3},
  pages   = {455--480},
  year    = {2023},
  note    = {Hebrew},
}

@article{yisraeli2024meditative,
  author  = {Yisraeli, Oded},
  title   = {`{As} If {I} Saw Them with My Eyes': The Meditative Kabbalah of {Rabbi Shem Tov Ibn Gaon}},
  journal = {Jewish Studies},
  volume  = {59},
  number  = {2},
  pages   = {39--65},
  year    = {2024},
  note    = {Hebrew},
}

@article{liu2024lostmiddle,
  author    = {Liu, Nelson F. and Lin, Kevin and Hewitt, John and Paranjape, Ashwin and Bevilacqua, Michele and Petroni, Fabio and Liang, Percy},
  title     = {Lost in the Middle: How Language Models Use Long Contexts},
  journal   = {Transactions of the Association for Computational Linguistics},
  volume    = {12},
  pages     = {157--173},
  year      = {2024},
  publisher = {MIT Press},
  doi       = {10.1162/tacl_a_00638},
}

@article{yisraeliForthcomingTradition,
  author  = {Yisraeli, Oded},
  title   = {Tradition and Innovation in Fourteenth-Century Supercommentaries on {Nahmanides'} Kabbalistic Interpretations of the {Torah}},
  journal = {Shnaton: An Annual for Biblical and Ancient Near Eastern Studies},
  year    = {forthcoming},
  note    = {Hebrew; translated title},
}

@book{idel1998recanati,
  author    = {Idel, Moshe},
  title     = {{R. Menahem Recanati} the Kabbalist},
  volume    = {1},
  address   = {Jerusalem and Tel Aviv},
  publisher = {Schocken},
  year      = {1998},
  note      = {Hebrew; translated title},
}

@article{idel1979unknown,
  author  = {Idel, Moshe},
  title   = {An Unknown Commentary on {Nachmanides'} Mystic Doctrines},
  journal = {Da'at},
  number  = {2--3},
  pages   = {121--126},
  year    = {1979},
  note    = {Hebrew; issue dated 1978--1979},
}

@article{abrams1996orality,
  author  = {Abrams, Daniel},
  title   = {Orality in the Kabbalistic School of {Nahmanides}: Preserving and Interpreting Esoteric Traditions and Texts},
  journal = {Jewish Studies Quarterly},
  volume  = {3},
  number  = {1},
  pages   = {85--102},
  year    = {1996},
}

@article{lawee2017genre,
  author  = {Lawee, Eric},
  title   = {A Genre Is Born: The Genesis, Dynamics, and Role of {Hebrew} Exegetical Supercommentaries},
  journal = {Revue des {\'e}tudes juives},
  volume  = {176},
  number  = {3--4},
  pages   = {295--332},
  year    = {2017},
  doi     = {10.2143/REJ.176.3.3268994},
}

\end{document}